\documentclass[11pt,a4paper]{article}
\usepackage[hyperref]{acl2020}
\usepackage{times}
\usepackage{latexsym}

\usepackage{booktabs}
\usepackage{graphicx}
\usepackage{subfigure}
\usepackage[]{linguex}
\usepackage{CJKutf8}
\usepackage{listings}
\usepackage{multirow}

\usepackage{microtype}

\aclfinalcopy 
\def\aclpaperid{65} 

\title{Lessons from Computational Modelling of Reference Production in Mandarin and English}

\author{Guanyi Chen \and Kees van Deemter \\
  Department of Information and Computing Sciences \\
  Utrecht University \\
  \texttt{\{g.chen, c.j.vandeemter\}@uu.nl}   } 

\date{}

\begin{document}
\maketitle
\begin{abstract}
Referring expression generation (REG) algorithms offer computational models of the production of referring expressions. In earlier work, a corpus of referring expressions (REs) in Mandarin was introduced. In the present paper, we annotate this corpus, evaluate classic REG algorithms on it, and compare the results with earlier results on the evaluation of REG for English referring expressions. Next, we offer an in-depth analysis of the corpus, focusing on issues that arise from the grammar of Mandarin. We discuss shortcomings of previous REG evaluations that came to light during our investigation and we highlight some surprising results. Perhaps most strikingly, we found a much higher proportion of under-specified expressions than previous studies had suggested, not just in Mandarin but in English as well.
\end{abstract}

\begin{CJK}{UTF8}{gbsn}

\section{Introduction} \label{sec:intro}

\begin{figure*}[tbp]
    \centering
    \subfigure[]{
        \includegraphics[width=.35\textwidth]{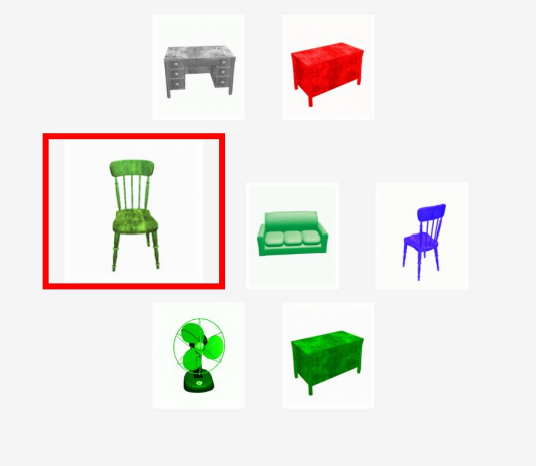}
    }
    \hspace*{0.5cm}
  \subfigure[]{
        \includegraphics[width=.35\textwidth]{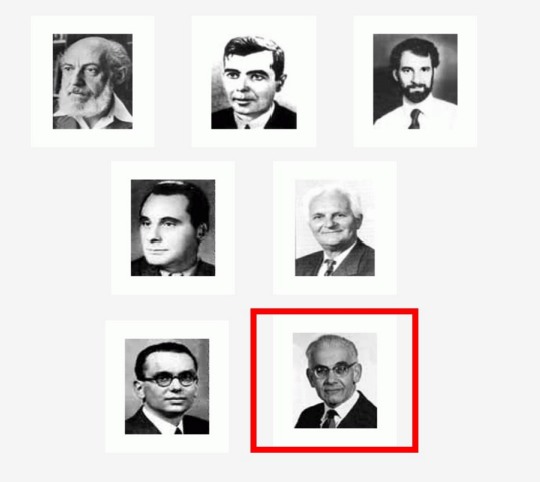}
    }
    \caption{Two scenes from the \textsc{tuna} experiment, in which (a) is a situation from the furniture domain while (b) is from the people domain.}
    \label{fig:tuna}
  \end{figure*}

Referring expression generation (REG) originated as a sub-task of traditional natural language generation systems~\citep[NLG,][]{reiter2000building}. The task is to generate expressions that help hearers to identify the referent that a speaker is thinking about. REG has important practical value in natural language generation \citep{gatt2018survey}, computer vision~\citep{mao2016generation}, and robotics~\citep{fang2015embodied}. Additionally, REG algorithms can be seen as models of human language use~\citep{van2016computational}.

In line with this second angle, and unlike REG studies which have started to use black-box Neural Network based models (e.g., \citet{mao2016generation, ferreira2018neuralreg} and \citet{cao2019referring}), we focus on two aspects (cf., \citet{krahmer2012computational}): 1) designing and conducting controlled elicitation experiments, yielding corpora which are then used for analysing and evaluating REG algorithms to gain insight into linguistic phenomena, e.g., \textsc{gre3d3}~\citep{dale2009referring}, \textsc{tuna}~\citep{gatt2007evaluating,deemter2012generation}, \textsc{coconut}~\citep{jordan2005learning}, and \textsc{MapTask}~\citep{gupta2005automatic}. 2) designing algorithms that mimic certain behaviours used by human beings, for example the maximisation of discriminatory power~\citep{dale1989cooking}  and/or the preferential use of cognitively ``attractive'' attributes~\citep{dale1995computational}; see \citet{gatt2013production} for discussion.

The focus of these studies was mostly on Indo-European languages, such as English, Dutch~\citep{koolen2010d} and German~\citep{howcroft2017g}. Recently researchers have started to have a look at Mandarin Chinese~\citep{van-deemter-etal-2017-investigating}, collecting a corpus of Mandarin REs, namely \textsc{mtuna}.
So far, only a preliminary analysis has been performed on \textsc{mtuna}, and this analysis has focused on issues of Linguistic Realisation ~\citep{van-deemter-etal-2017-investigating}: the REs in the corpus have not yet been compared with those in other languages, and the performance of REG algorithms on the corpus has not been evaluated. 

To fill this gap, we provide a more detailed analysis of the use of Mandarin REs on the basis of the \textsc{mtuna} corpus. We annotated the \textsc{mtuna} corpus in line with the annotation scheme of \textsc{tuna}~\citep{van2006manual}, after which we used this annotation to evaluate the classical REG algorithms and compared the results with those for the English \textsc{etuna} corpus. Since it has been claimed that Mandarin favours brevity over clarity
-- the idea that Mandarin is ``cooler" than these other languages ~\citep{newnham1971about,huang1984distribution} -- relying more on communicative context for disambiguation than western languages, we concentrated on the use of over- and under-specification. After all, if Mandarin favours brevity over clarity to a greater extent than English and Dutch, then one would expect to see less over-specification and more over-specification in Mandarin.
\section{Background}

The analysis reported in paper is based on the \textsc{mtuna} (for Mandarin) and \textsc{etuna} (for English) corpus. We start by briefly introducing the \textsc{tuna} experiments in general, and we highlight some special features of \textsc{mtuna} together with its initial findings.

\subsection{The \textsc{tuna} Experiments}

\textsc{tuna}~\citep{gatt2007evaluating, van2007evaluating} is a series of controlled elicitation experiments that were set up to aid computational linguist's understanding of human reference production. In particular, the corpora to which these experiments gave rise were employed to evaluate REG algorithms, by comparing their output with the REs in these corpora. The stimuli in the \textsc{tuna} experiments were divided into two types of visual scenes: scenes that depict furniture and scenes that depict people. Figure~\ref{fig:tuna} shows an example for each of these two types of scenes. In each trial, one or two objects in the scene were chosen as the target referent(s), demarcated by red borders. The subjects were asked to produce referring expressions that identify the target referents from the other objects in the scene (their ``distractors"). For example, for the scene in Figure~\ref{fig:tuna}, one might say \emph{the large chair}. 
The trials in the people domain were intended to be more challenging than those in the furniture domain.

The resulting corpus, which we will call \textsc{etuna}, was subsequently studied for evaluating a set of ``classic'' REG algorithms~\citep{deemter2012generation}.  Although RE has given rise to a good number of other corpora, with subtly different qualities (e.g., \citet{dale2009referring}), we focus here on the \textsc{tuna} corpora for two reasons: firstly the \textsc{etuna} corpus was used in a series of Shared Task Evaluation Campaign~\citep{gatt2010introducing}, which caused it to be relatively well known. Secondly and more importantly from the perspective of the present paper, \textsc{etuna} inspired a number of similarly constructed corpora for Dutch~\citep[\textsc{dtuna},][]{koolen2010d}, German~\citep[\textsc{gtuna},][]{howcroft2017g}, and Mandarin~\citep{van-deemter-etal-2017-investigating}.

\subsection{The Mandarin \textsc{tuna}} \label{sec:mtuna}

The different \textsc{tuna} corpora were set up in highly similar fashion: for instance, they all use a few dozen stimuli, which were offered in isolation (i.e., participants were encouraged to disregard previous scenes and previous utterances), and chosen from the same sets of furniture and people images; furthermore, participants were asked to enter a type-written RE following a question. 

Yet there were subtle differences between these corpora as well, reflecting specific research questions that the various sets of authors brought to the task. The stimuli used by \textsc{mtuna} were inherited from the \textsc{dtuna}, where there are totally 40 trials. Different from other \textsc{tuna}s which always asked subjects essentially the same question, namely \emph{Which object/objects appears/appear in a red window?}, \textsc{mtuna}
distinguished between referring expressions in subject and object position.\footnote{This was done because the literature on Mandarin (e.g., \citet{chao1965grammar}) suggests that Mandarin NPs in pre-verbal position may be interpreted as definite unless there is information to the contrary.}
More precisely, subjects were asked to use REs for filling in blanks in either of the following patterns: 
\a.[(a)] \_\_\_\_ 在红色方块中
\glt `Please complete the sentence: \_\_\_\_ is in the red frame(s)'
\b.[(b)] 红色方块中的是 \_\_\_\_ 
\glt `What’s in the red frame is \_\_\_\_'
    
where (a) asked subjects to place the referring expression in subject position while (b) asked to place it in object position. The initial analysis in \citet{van-deemter-etal-2017-investigating} focused on how definiteness was expressed in Mandarin REs. They found that 1) most definite REs are bare nouns. 2) indefinite REs also appear quite often, especially in subject position. 
\section{Research Questions} \label{sec:rq}

Analogous to studies of earlier \textsc{tuna} corpora, our primary research question (\textbf{RQ1}) is 
\emph{how classic REG algorithms perform on \textsc{mtuna} and how this is different from the performance on \textsc{etuna}?} We were curious to see whether the value of each evaluation metric for each algorithms will change very much, and whether the rank order of the algorithms stays the same. If, as 
hypothesised, Mandarin prefers brevity over clarity, then the Full Brevity algorithm (which always yields REs with minimally number of properties), is expected to have higher performance on \textsc{mtuna} than on \textsc{etuna}. The expected effect on other classic algorithms is less clear. 

It is thought that, since \texttt{TYPE} helps create a ``conceptual gestalt'' of the target referent (which benefits the hearer~\citep[Chapter 4]{levelt1993speaking}) 
speakers tend to include a \texttt{TYPE} in their REs regardless of its discriminatory power.\footnote{Note that 92.25\% of the REs in \textsc{etuna} contain a superfluous \texttt{TYPE}~\citep{van2007evaluating}).} For this reason, algorithms such as the Incremental Algorithm~\citep{dale1995computational} always append a \texttt{TYPE} to the REs they produce. However, 
\citet{lv1979problems} found that the head of a noun phrase in Mandarin is often omitted if this noun is the only possibility given the context. This suggests that, if all objects in a scene share the same \texttt{type} (e.g., all the objects in the people domain of \textsc{tuna} are male scientists), then it is less likely for Mandarin speakers to express a \texttt{TYPE}. Accordingly, our second research question (\textbf{RQ2}) asks \emph{to what extent the role of \texttt{TYPE} differs between English and Mandarin.} Connected with this, we were curious to what extent this issue affects the performance of the classic REG algorithms. 

As discussed in section~\ref{sec:intro}, the coolness hypothesis stated that Chinese relies more on the communicative context for disambiguation than western languages, such as English, based on which Chinese is also seen as a discourse-based language while English is a sentence-based language.
The existence of primary evidence for this issue in REG was identified in~\citet{van-deemter-etal-2017-investigating}, indicating that Mandarin speakers rarely explicitly express number, maximality and giveness in REs, and in~\citet{chen2018modelling}, indicting that they sometimes even drop REs.
In this study, we were curious about (\textbf{RQ3}) \emph{the use of over-specification and under-specifications in \textsc{mtuna} versus \textsc{etuna}}, hypothesising that Mandarin REs use fewer over-specifications and more under-specifications than English. 

We have seen that \textsc{mtuna} asked its participants to produce REs in different syntactic positions. 
\citet{van-deemter-etal-2017-investigating} found more indefinite NPs in the subject position, which is inconsistent with linguistic theories~\citep{james2009syntax} that suggests subjects and other pre-verbal positions favour definiteness. Building on these findings, we investigated (\textbf{RQ4}) \emph{how syntactic position 
influences the use of over-/under-specification and the performance of REG algorithms}.

\section{Method}

\begin{table*}[t]
\centering
\small
\begin{tabular}{llcccccccc}
\toprule
                  & Domain & \textbf{Total} & \textbf{Mini.} & \textbf{Real} & \textbf{Nom.} & \textbf{Num.} & \textbf{Wrong} & \textbf{Other} & \textbf{Under}\\\midrule
\multirow{2}{*}{\textsc{mtuna}} & furniture & 377 & 46 & 117 & 132 & 2 & 11 & 5 & 64\\
                 & people & 371 & 16 & 216 & 68 & 13 & 4 & 6 & 48 \\\midrule
\multirow{2}{*}{\textsc{mtuna-ol}} & furniture & 264 & 9 & 83 & 104 & 0 & 8 & 4 & 56\\
& people & 222 & 14 & 144 & 36 & 2 & 1 & 3 & 22 \\\midrule
\multirow{2}{*}{\textsc{etuna}} & furniture & 158 & 1 & 58 & 62 & 0 & 0 & 0 & 37 \\
                  & people & 132 & 3 & 75 & 37 & 0 & 0 & 0 & 7 \\
\bottomrule
\end{tabular}
\caption{Frequencies of referring expressions that fall in each type specifications in \textsc{mtuna}, \textsc{mtuna-ol} and \textsc{etuna} respectively. Specifically, \textbf{total} is the total number of descriptions in each corpus. \textbf{mini.} is the minimal over-specification, \textbf{real} is the real over-specification, \textbf{nom.} is the nominal over-specification, \textbf{num.} is the numerical over-specification, \textbf{wrong} is the duplicate-attribute over-specification, \textbf{other} stands for the RE that cannot be classified into any of these categories, and \textbf{under} is the under-specification.}
\label{tab:over}
\end{table*}

\begin{table*}[tbp]
\centering
\small
\begin{tabular}{lcclcc}
\toprule
 \multicolumn{3}{c}{\textsc{Furniture}} & \multicolumn{3}{c}{\textsc{People}}\\
 \cmidrule(lr){1-3}\cmidrule(lr){4-6}
 Model & DICE (SD) & PRP & Model & DICE (SD) & PRP \\
 \midrule 
 IA-COS & \textbf{0.875 (0.17)} & \textbf{55.7} & IA-GBHOATSS & 0.637 (0.26) & 16.3 \\
 IA-CSO & 0.847 (0.21) & 55.1 & IA-BGHOATSS & 0.629 (0.25) & 15.5\\
 IA-OCS & 0.797 (0.16) & 20.5 & IA-GHBOATSS & 0.617 (0.25) & 13.0 \\
 IA-SCO & 0.754 (0.18) & 15.0 & IA-BHGOATSS & 0.577 (0.24) & 7.5 \\
 IA-OSC & 0.740 (0.20) & 18.3 & IA-HGBOATSS & 0.589 (0.23) & 6.1 \\
 IA-SOC & 0.690 (0.21) & 14.7 & IA-HBGOATSS & 0.559 (0.24) & 6.1 \\
 - & - & - & IA-SSTAOHBG & 0.347 (0.23) & 1.9 \\
 \midrule
 FB+\texttt{TYPE} & 0.830 (0.18) & 39.9 & FB+\texttt{TYPE} & \textbf{0.669 (0.26)} & \textbf{23.2} \\
FB & 0.574 (0.25) & 3.0 & FB & 0.446 (0.32) & 9.9 \\
 GR & 0.802 (0.21) & 39.3 & GR & 0.613 (0.29) & 19.9 \\
 \bottomrule
\end{tabular}
\caption{Experiment results on \textsc{mtuna}, in which the string after each IA algorithm represents the preference order it uses. For example, ``COS'' means \texttt{COLOUR} $>$ \texttt{ORIENTATION} $>$ \texttt{SIZE} and ``BGHOATSS'' stands for \texttt{hasGlasses} $>$ \texttt{BEARD} $>$ \texttt{HAIR} $>$ \texttt{ORIENTATION} $>$ \texttt{AGE} $>$ \texttt{hasTie} $>$ \texttt{hasShirt} $>$ \texttt{hasSuit}.}
\label{tab:mtuna}
\end{table*}

Before we address the four research questions in section~\ref{sec:rq}, we explain how we annotated the corpus. 
The annotated corpus is available at \url{github.com/a-quei/mtuna-annotated}

\subsection{Annotating the Corpus} \label{sec:anno_tuna}

1650 REs were semantically annotated (after omitting some unfinished REs from the corpus) following the scheme of~\citet{van2006manual}.\footnote{This includes the trials that have one target referent and those that have two targets, but, in this paper, we focus on the former one. The annotated corpus is public available at: xxx.} For simplicity, instead of XML we use the JSON for the annotation. Because the scenes stay the same when different subjects accomplished the experiment, we annotated the scene and the REs in \textsc{mtuna} separately. 
For the attribute \texttt{hairColour}, both ~\cite{van2006manual} and~\citet{gatt2008xml} (and all the annotate scheme used by the previous \textsc{tuna} corpora) annotated both hair colour and beard colour as \texttt{hairColour}. 
However, this would cause us to 
overlook some key phenomena, because some participants used the colour of a person's beard for distinguishing the target. Therefore, we decided to use \texttt{hairColour} and \texttt{beardColour} as separate attributes. 
As pointed out in~\citet{deemter2012generation}, since the attribute \texttt{hairColour} is depend on \texttt{hasHair}, the authors merged these two into a single attribute \texttt{Hair} during the evaluation. We did the same thing and obtained two merged attributes: \texttt{Hair} and \texttt{Beard}.

To avoid compromising the comparison between 
\textsc{mtuna} and \textsc{etuna}, 
we did not only annotate \textsc{mtuna} but we also re-annotated the \textsc{etuna} corpus, using the same annotators. Details about which properties were annotated and examples of annotated REs can be found in Appendix A.

\subsection{Annotating Over-/Under-specifications}

To gain an insightful analysis of the speakers' use of over- and under-specification, and to ensure that our annotations are well defined,
%
we will offer some definitions. In addition, 
given our interest in the role of \texttt{TYPE}, we will sub-categorise by distinguishing different types of over-specifications. Concretely, we asked the annotators to consider the following types of specifications:

\noindent\textbf{Minimal Description.}~~an RE that successfully singles out the target referent and does this by using the minimum possible number of properties. These are the REs that match \citeauthor{dale1995computational}'s Full Brevity;

\noindent\textbf{Numerical Over-specification.}~~an RE that uses more properties than the corresponding minimal description uses, yet the removal of any of them results in a referential confusion. For instance, for the scene in~\ref{fig:tuna}(a), the RE \emph{the green chair} is a numerical over-specification as it uses more properties than the minimal description \emph{the large one};

\noindent\textbf{Nominal Over-specification.}~~an RE from which only one of its properties is removable, namely the \texttt{TYPE} of the target;

\noindent\textbf{Real Over-specification.}~~an RE from which at least one of its non-\texttt{TYPE} properties is removable;

\noindent\textbf{Under-specification.}~~an RE all of whose properties are true of the referent but 
that causes referential confusion (i.e., it is not a distinguishing description in the sense of \citet{dale1992generating}); 

\noindent\textbf{Wrong Description.}~~an RE whose properties use one or more incorrect values for a given attribute. In line with previous TUNA evaluations, we only consider a value to be wrong if it could prevent a hearer from recognising the target. For example, the RE \emph{the pink chair} is not called wrong if the referent is a red chair. 

We annotated each RE in both corpora \footnote{When applying this annotation scheme to REs that have multiple targets, adaptations need to be made. But since the focus of this paper is on singular REs, we will not offer details.}, and we annotated each scene in each corpus. Thus, for each RE, we annotate which of the above specification types
it falls in, and how many over-specified/under-specified properties the RE contains.
In Appendix B, Table 7 records, for each scene, how many different minimal descriptions the scene permits (most often just 1, but sometimes 2 or 3).  
The results per RE are depicted in Table~\ref{tab:over}\footnote{We observed a large number of minimal descriptions in the furniture domain of \textsc{mtuna}. This is a result of the fact that some trials in \textsc{mtuna} use \texttt{TYPE} in their minimal descriptions. 
} and the results per scene are in Appendix B.
\section{Analysis} \label{sec:analysis}

\begin{table*}[t]
\centering
\small
\begin{tabular}{lcccclcccc}
\toprule
  \multicolumn{5}{c}{{\textsc{Furniture}}} & \multicolumn{5}{c}{{\textsc{People}}}\\
  \cmidrule(lr){1-5} \cmidrule(lr){6-10}
  & \multicolumn{2}{c}{\textsc{etuna}} & \multicolumn{2}{c}{\textsc{mtuna-ol}} &  & \multicolumn{2}{c}{\textsc{etuna}} & \multicolumn{2}{c}{\textsc{mtuna-ol}} \\
  \cmidrule(lr){2-3} \cmidrule(lr){4-5} \cmidrule(lr){7-8}\cmidrule(lr){9-10}
 Model & DICE (SD) & PRP & DICE (SD) & PRP & Model & DICE (SD) & PRP & DICE (SD) & PRP \\
 \midrule
 IA-COS & \textbf{0.919 (0.12)} & \textbf{62.8} & \textbf{0.915 (0.14)} & \textbf{65.5} & IA-GBHOATSS & \textbf{0.862 (0.17)} & 50.0 & 0.724 (0.22) & 22.8 \\
 IA-CSO & \textbf{0.919 (0.12)} & \textbf{62.8} & \textbf{0.915 (0.14)} & \textbf{65.5} & IA-BGHOATSS & 0.861 (0.17) & \textbf{50.8} & 0.719 (0.21) & 21.0 \\
 IA-OCS & 0.832 (0.14) & 26.3 & 0.823 (0.15) & 25.4 & IA-GHBOATSS & 0.774 (0.20) & 27.3 & 0.674 (0.25) & 19.6 \\
 IA-SCO & 0.817 (0.14) & 20.5 & 0.808 (0.15) & 19.4 & IA-BHGOATSS & 0.761 (0.19) & 25.0 & 0.621 (0.22) & 7.8 \\
 IA-OSC & 0.805 (0.16) & 23.7 & 0.798 (0.17) & 23.8 & IA-HGBOATSS & 0.705 (0.17) & 3.8 & 0.609 (0.22) & 4.1 \\
 IA-SOC & 0.782 (0.16) & 19.9 & 0.767 (0.17) & 19.4 & IA-HBGOATSS & 0.670 (0.19) & 4.5 & 0.570 (0.23) & 3.7 \\
 - & - & - & - & -                                & IA-SSTAOHBG & 0.339 (0.10) & 0.0 & 0.285 (0.17) & 0.0 \\
 \midrule
 FB+\texttt{TYPE} & 0.849 (0.17) & 41.7 & 0.849 (0.16) & 42.5 &  FB+\texttt{TYPE} & 0.847 (0.17)  & 44.7 & \textbf{0.734 (0.23)} & \textbf{27.4} \\
 FB & 0.590 (0.23) & 0.6 & 0.602 (0.24) & 3.6 &  FB & 0.556 (0.16) & 2.3 & 0.541 (0.26) & 11.0 \\
 GR & 0.849 (0.17) & 41.7 & 0.849 (0.16) & 42.5 & GR & 0.727 (0.25) & 33.3 & 0.650 (0.28)& 21.9 \\
 \bottomrule
\end{tabular}
\caption{Experiment results on the \textsc{mtuna-ol} and \textsc{etuna}. Algorithms are listed from top to bottom in order of their performance on \textsc{etuna}.}
\label{tab:mtuna_etuna}
\end{table*}

Before reporting results and analysis,
we explain what datasets and algorithms were analysed, and how evaluation was performed.  


\noindent \textbf{Dataset.} The sources of our dataset are the \textsc{mtuna} and \textsc{etuna} corpora. For Mandarin, we used the whole \textsc{mtuna} dataset. For comparing between languages fairly, we only used REs for scenes that were shared between \textsc{mtuna} and \textsc{etuna}; we call this set of shared scenes \textsc{mtuna-ol}. The original \textsc{mtuna} has 20 trials, with 10 trials for each domain. The \textsc{mtuna-ol} and \textsc{etuna} contains 13 trials, in which there are 7 and 6 trials from furniture and people domain respectively. (More details of which scene is used can be found in the Appendix.) 

\noindent \textbf{Algorithms.} We tested the classic REG algorithms, including: 1) the Full Brevity algorithm~\citep[FB][]{dale1989cooking}: an algorithm that finds the shortest RE; 2) the Greedy algorithm~\citep[GR][]{dale1989cooking}: an algorithm that iteratively selects properties that rule out a maximum number of distractors (i.e., a property that has the highest ``Discriminative Power''); and 3) the Incremental Algorithm: an algorithm that makes use of a fixed ``preference order" of attributes~\citep[IA][] {dale1995computational}.

\noindent \textbf{Evaluation Metrics.} We used what are still the most commonly used metrics for evaluating attribute choice in REG. One is the DICE metric~\citep{dice1945measures}, which measures the overlap between two attributes sets:
$$\mbox{DICE}(\mathcal{D}_H, \mathcal{D}_A) = \frac{2 \times |\mathcal{D}_H \cap \mathcal{D}_A|}{|\mathcal{D}_H| + |\mathcal{D}_A|}$$
where $\mathcal{D}_H$ is the set of attributes expressed in the description produced by a human author and $\mathcal{D}_A$ is the set of attributes expressed in the logical form generated by an algorithm. We also report the ``perfect recall percentage'' (PRP), the proportion of times the algorithm achieves a DICE score of 1, which is seen as a indicator of the recall of an algorithm.

\subsection{Performance of Algorithms on \textsc{mtuna}}



We report the evaluation results on \textsc{mtuna} and \textsc{mtuna-ol} in the Table~\ref{tab:mtuna} and~\ref{tab:mtuna_etuna}. For the FB algorithm, we tested both the original version and the version that always appends a \texttt{TYPE} (named FB+\texttt{TYPE}). Moreover, since we did not observe any significant difference in the frequencies of use of each attribute between \textsc{mtuna} and \textsc{etuna} corpora, 
we let the IA make use of the same set of preference orders as~\citet{deemter2012generation}.

In line with the previous findings in other languages, in the furniture domain, it is IA (with a good preference order) that perform the best in both \textsc{mtuna} and \textsc{mtuna-ol}. Interestingly, the people domain yields very different results: this time, FB+\texttt{TYPE} becomes the winner. 

An ANOVA test comparing GR, FB+\texttt{TYPE}, and the best IA suggests a significant effect of algorithms on both domains and on both \textsc{mtuna} and \textsc{mtuna-ol}
(Furniture: $F(2, 1008)=49.20, p=.002$; People: $F(2, 1065)=11.97, p<.001$) and \textsc{mtuna-ol} (Furniture: $F(2, 699)=14, p<.001$; People: $F(2, 622)=4.22, p=.015$). As for each algorithm, by Tukey's Honestly Significant Differences (HSD), we found that IA defeats other algorithms in the furniture domain in both corpora ($p<.001$) and that the victory of FB+\texttt{TYPE} for people domain is significant in \textsc{mtuna} ($p=.001$) but not in \textsc{mtuna-ol} ($p=0.96$).


The scores for algorithms in the people domain are much lower than those in the furniture domain, even lower than the scores for the people domain in \textsc{etuna}.
This may be because, based on the numbers in Table~\ref{tab:over}, a Chi Squared Test suggests that, in \textsc{mtuna}, there are more real over-specifications ($\chi^2(1, 747)=55.95, p < .001$) but fewer nominal over-specifications ($\chi^2(1, 747)=26.57, p < .001$) in the people domain than in the furniture domain\footnote{This highlights the importance of sub-categorising the different kinds of over-specifications, as we have done in section 4}. As for the former, real over-specifications are notoriously hard to model accurately by deterministic REG algorithms, which is one of the motivations behind probabilistic modelling~\citep{van2019conceptualization} or Bayesian Modelling~\citep{degen2020redundancy}; such an approach might have additional benefits for the modelling of reference in Mandarin. The relative lack of nominal over-specifications in Mandarin descriptions of people 
could be addressed along similar lines, adding \texttt{TYPE} probabilistically. Another evidence
is that, in the \textsc{mtuna} people domain, FB outperforms many IAs on PRP, which does not happen in the furniture domain.

By comparing the results for \textsc{mtuna} and \textsc{mtuna-ol}, we found that the rank order (by performance) of algorithms stays the same, but the absolute scores for the latter corpus are much higher. If we look into the annotations for the trials from \textsc{mtuna} that are not in \textsc{mtuna-ol} (Appendix B), most of these trials have multiple possible minimal descriptions and numerical over-specifications. Every RE in the corpus that results in a successful communication 
can be seen as either a minimal description or a numerical over-specification, with 0 or more attributes added to it.
When computing the DICE similarity score between a generated RE and human produced REs, if it is close to a minimal description, it will differ from another minimal description. 
For example, suppose we have a trial having two miminal descriptions: \emph{the large one} and \emph{the green one}. Our FB produce the second minimal description (as it can only produce one RE at a time). When we computing DICE, we obtain $\frac{2}{3}$ for the RE \emph{the green chair} while 0 for the RE \emph{the large chair}, but, in fact, either of them has only one superfluous attributes.
This implies that when a corpus contains multiple minimal REs, this will artificially lower the DICE scores.\footnote{An analogous problem has been identified in the task of evaluating image capturing~\citep{yi-etal-2020-improving}, where the collision of multiple references for a single image was considered.} For the same reason, the performance of FB increases a lot from \textsc{mtuna}/People to \textsc{mtuna-ol}/People 
because all trials in \textsc{mtuna-ol} have only one possible minimal description. Another reason lies in the decrease in the number of under-specifications from \textsc{mtuna}/People to \textsc{mtuna-ol}/People.

\subsection{Cross-linguistic Comparison} \label{sec:compare}

Table~\ref{tab:mtuna_etuna} reports the results for both \textsc{mtuna-ol} and \textsc{etuna}, from which, except for the fact that FB+\texttt{TYPE} becomes having the best performance, we see no difference on the order of the their performance. 
An interesting observation is that, after correcting a few errors in the annotation of \textsc{etuna} (cf. section~\ref{sec:anno_tuna}), the difference between IA and FB+\texttt{TYPE} is no longer significant in the people domain in terms of Tukey's HSD (compare the conclusion in~\citet{deemter2012generation}). In other words, in both languages there is no significant difference between the performance of these two algorithms on the people domain. We also checked the influence of language on the performance of FB and FB+\texttt{TYPE}: the influence of the former is significant ($F(1, 349)=23.63, p<.001$) while that of later is not ($F(1, 349)=0.36, p=.548$). This suggest that, in fact, it is English speakers who show more brevity, except in terms of use of \texttt{TYPE}. This might also explain the differences in absolute scores for all algorithms in both \textsc{etuna} and \textsc{mtuna}, 
especially in the people domain. Another possible reason for these differences
is the fact that the REs in \textsc{mtuna-ol} show slightly higher diversity in the choice of content than \textsc{etuna}, as the standard deviations for every model is higher.

\subsection{RQ2: the Role of \texttt{TYPE}} \label{sec:type}

\begin{figure}
    \centering
    \includegraphics[scale=0.55]{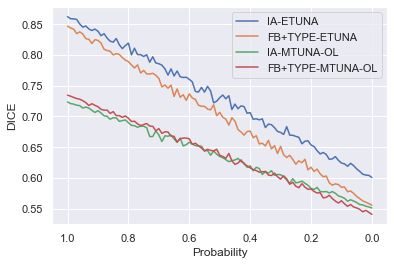}
    \caption{Change of the performance with respected to different probabilities of inserting superfluous \texttt{TYPE} for either the FB+\texttt{TYPE} and IA on either the people domain of \textsc{mtuna-ol} and \textsc{etuna}.}
    \label{fig:prob}
\end{figure}

On the use of \texttt{TYPE}, we first look at the number of REs that uses \texttt{TYPE} in \textsc{mtuna-ol} and \textsc{etuna}. 98.4\% and 95.93\% of REs in the furniture and people domains of \textsc{etuna} contain \texttt{TYPE}. For \textsc{mtuna-ol}, those numbers are 91.29\% and 74.77\%, suggesting that Mandarin speakers are less likely to use superfluous \texttt{TYPE}. Second, for \citeauthor{lv1979problems}'s hypothesis introduced in section~\ref{sec:rq}, we observed a smaller proportion of uses of \texttt{TYPE} in the people domain ($\chi^2(1, 485)=24.16, p < .001$), where all the objects share the same value of \texttt{TYPE}. Comparing the performance of REG algorithms on the furniture domain of \textsc{mtuna} and \textsc{mtuna-ol}, the difference is not as huge as that in the people domain. 
This implies that the complement of \citeauthor{lv1979problems}'s hypothesis might also hold, namely, if the value of \texttt{TYPE} is {\em not} the only possibility, then it will not be omitted. 

To further assess the role of \texttt{TYPE} and to find more evidence regarding \citeauthor{lv1979problems}'s hypothesis, we investigated how introducing uncertainties in whether or not to include a \texttt{TYPE} affects the performance of REG algorithms for the people domain. We tried out different probabilities, and for each probability for inserting the \texttt{TYPE} we ran the algorithm 100 times; we report the average DICE score, drawing the lines indicating the change of performance over different probabilities in Figure~\ref{fig:prob}.

We found that: 1) the decrease in performance on \textsc{mtuna-ol} is smaller than that on \textsc{etuna}; 2）IA and FB+\texttt{TYPE} have similar performance for Mandarin while IA performs better for English; 3) The difference between the performance of these algorithms 
becomes smaller when the influence of \texttt{TYPE} is ignored (i.e., when the probability of inserting \texttt{TYPE} is close to zero), especially for the Full Brevity algorithm. On top of these findings, we observe that although Mandarin speakers are less likely to use superfluous \texttt{TYPE}, always adding \texttt{TYPE} achieves the best performances for all the algorithms. Such a result maybe be caused by the dependencies between the use of different properties. In other words, introducing uncertainty to only the \texttt{TYPE} cannot sufficiently model the uncertainties in REG: when to drop a \texttt{TYPE} might also depend on the use of other properties.

\subsection{RQ3: Over-/Under-specification}

In light of Table~\ref{tab:over}, some obvious conclusions can be drawn. For example, more ``real" over-specifications are used for more complex domains (i.e., the people domain) than for simple ones. Focusing on RQ3 in section~\ref{sec:rq}, its two hypotheses are both rejected: no significant difference has been found in the use of over-specifications ($\chi^2(1, 775)=0.82, p = 0.052$) or in the use of under-specifications ($\chi^2(1, 775)=0.745, p =0.105$). Focusing on the people domain, where FB+\texttt{TYPE} performed better in English than in Mandarin, 
we found no significant difference ($\chi^2(1, 354)=2.53, p =0.112$).

\subsection{RQ4: Syntactic Position}

\begin{table}[t]
\centering
\small
\begin{tabular}{lcc}
\toprule
 Model & Furniture & People \\
 \midrule
 IA (subj.) & 0.940 (0.11)$^\dag$ & 0.728 (0.23) \\
 IA (obj.) & 0.890 (0.16) & 0.719 (0.21) \\ \midrule
 GR (subj.) & 0.884 (0.13)$^\dag$ & 0.629 (0.30) \\
 GR (obj.) & 0.815 (0.18) & 0.669 (0.25) \\\midrule
 FB+\texttt{TYPE} (subj.) & 0.884 (0.13)$^\dag$ & 0.736 (0.23) \\
 FB+\texttt{TYPE} (obj.) & 0.815 (0.18) & 0.733 (0.22) \\
 \bottomrule
\end{tabular}
\caption{The performance of REG algorithms for REs in different syntactic positions, in which IA is the IA with highest performance in the previous experiments, i.e., the IA-COS and IA-GBHOATSS. $\dag$ indicates that there is significant influence of the syntactic position on that algorithm in that domain.}
\label{tab:position}
\end{table}

For RQ4, we counted the number of real over-specifications and under-specification in subject and object position. In the \textsc{mtuna-ol} corpus, there are 247 and 239 descriptions in the subject and object positions, respectively. No significant difference on the use of over-specifications was found ($\chi^2(1, 485)=1.57, p =0.209$) but a significant difference regarding the use of under-specifications did exist ($\chi^2(1, 485)=19.27, p < .001$). Considering the fact that there are more indefinite RE in subject position~\citep{van-deemter-etal-2017-investigating}, the present finding might suggest that those indefinite REs are not suitable for identifying a target referent. It appears that further research is required to understand these issues in more detail.

As for the computational modelling, generally speaking, all algorithms performed better for REs in subject position than for REs is object position, with one exception, namely the GR algorithm for the people domain; the difference is significant in the Furniture domain, but is not in the people domain, possibly because the furniture domain contains more under-specifications. 
\section{Discussion}

\subsection{Lessons about RE use} 

Regarding the ``coolness" hypothesis, which focuses on the trade-off between brevity and clarity, we found that the brevity of Mandarin is only reflected in the use of \texttt{TYPE} but not in the other attributes, and, interestingly, no evidence was found that this leads to a loss of clarity; our findings are consistent with the possibility that Mandarin speakers may have found a better optimum than English. 

Although Mandarin speakers are less likely to over-specify \texttt{TYPE}, following ~\citet{lv1979problems},
we conclude that \texttt{TYPE} is often omitted \emph{if and only if} it has only one possible value given the domain. 
This appears to happen ``unpredictably'' (i.e., in one and the same situation, \texttt{TYPE} is often expressed but often omitted as well). However, we saw that introducing probability for the use of \texttt{TYPE} alone does not work well.  
This suggests that, to do justice to the data, a REG model may have to embrace non-determinism more wholeheartedly, as in the probabilistic approaches of~\citet{van2019conceptualization} and~\citet{degen2020redundancy}.

We found significant influence of the syntactic position of the RE on the use of under-specification and on the performance of REG algorithms. This flies in the face of earlier research on REG -- which has tended to ignore syntactic position -- yet it is in line with the theory of \citet{chao1965grammar}. On the other hand, it gives rise to various questions: {\em why} are more under-specifications used in subject positions, and why do all REG models perform better for REs in subject positions than for those in object position? These questions invite further studies including, for example, reader experiments to find out how REs in different positions are comprehended. It would also be interesting to investigate what role syntactic position plays in other languages, where this issue has not yet been investigated.

Perhaps our most surprising findings regard the use of under-specification: firstly (deviating from what ~\citet{van-deemter-etal-2017-investigating} hypothesised), we did not find significantly more under-specifications in \textsc{mtuna} than in \textsc{etuna}. 
We found a very substantial proportion, of nearly 20\%, under-specified REs in both  \textsc{mtuna} and \textsc{etuna}. This was surprising, because, at least in Western languages, in situations where Common Ground is unproblematic~\citep{horton1996speakers}, under-specification is widely regarded as a rarity in the language use of adults, to such an extent that existing REG algorithms are typically designed to prevent under-specification completely (see e.g.,~\citet{krahmer2012computational}). Proportions of under-specifications in corpora are often left reported, but ~\citep{koolen2011factors} report that only 5\% of REs in \textsc{dtuna} were under-specifications.\footnote{The difference might be that \textsc{dtuna} used participants who came into the lab separately, whereas \textsc{mtuna} participants sat together in a classroom.}

These findings give rise to the following questions: 1) Why did previous investigators either find far fewer under-specified REs (e.g., \citet{koolen2011factors}, see Footnote 8) or ignored under-specification? 
2) How does the presence of under-specification influence the performance of the classic REG algorithms (which never produce any under-specified REs, except when no distinguishing RE exists)? and 3) If a REG model aims -- as most do -- to produce human-like output, then what is the most effective way for them to model under-specification? 

\subsection{Lessons about REG Evaluation}

Most REG evaluations so far have made use of the DICE score~\citep{dice1945measures}.
However, on top of the discussions of \citet{Deemter2007Content} and of section~\ref{sec:analysis}, we identify the following three issues for evaluating REG with DICE. First, if a scene has multiple possible minimal descriptions or numerical over-specifications, then this causes DICE scores to be artificially lowered (section~\ref{sec:mtuna}) and hence distorted. Second, there is no guarantee that an RE with a high DICE score is a distinguishing description. Third, DICE punishes under-specification more heavily than over-specification. 
Suppose we have a reference RE $d$ which uses $n$ attributes, a over-specification $d_o$ with one more superfluous comparing to $d$ (so it uses $n+1$ attributes), and a under-specification $d_u$ which can be repaired to $d$ by adding one attribute (using $n-1$ attributes), the DICE score of $d_o$ is $2n / (2n+1)$ while $d_u$'s DICE is $2n - 2 / (2n-1)$. In other words, $d_o$ has a higher DICE than $d_u$. 
Whether this should be considered a shortcoming of DICE or a feature is a matter for debate.

Finally, our analysis suggests that previous \textsc{tuna} experiments may have been insufficiently controlled. For example, some trials in \textsc{mtuna} and \textsc{dtuna} use \texttt{TYPE} for distinguishing the target, causing nominal over-specifications not to be counted as over-specification. 
Different trials have different numbers of minimal descriptions and different numbers of numerical over-specifications. 
As shown in section~\ref{sec:analysis}, these issues impact evaluation results and this might cause the conclusions from evaluating algorithms with \textsc{tuna} not to be reproducable. 

Comparisons between corpora need to be approached with caution, and the present situation is no exception. For all the similarities between them, we have seen that there are significant differences in the ways in which the \textsc{tuna} corpora were set up.\footnote{Most \textsc{tuna} experiments involved type-written REGs, but \textsc{dtuna} elicited spoken REs. In most \textsc{tuna} experiments the linguistic context was uniform, but \textsc{mtuna} elicited REs in different syntactic positions, as we have seen.}
Although these differences exist for a reason (i.e., for testing linguistic hypotheses), we believe that it would be worthwhile to design new multilingual datasets, where care is taken to ensure that utterances in the different languages are elicited under circumstances that are truly as similar as they can be.
%

\section*{Acknowledgements}
We thank the anonymous reviewers for their helpful comments.
Guanyi Chen is supported by China Scholarship Council (No.201907720022).

\bibliography{acl2020}

\begin{thebibliography}{38}
\expandafter\ifx\csname natexlab\endcsname\relax\def\natexlab#1{#1}\fi

\bibitem[{Cao and Cheung(2019)}]{cao2019referring}
Meng Cao and Jackie Chi~Kit Cheung. 2019.
\newblock Referring expression generation using entity profiles.
\newblock \emph{arXiv preprint arXiv:1909.01528}.

\bibitem[{Chao(1965)}]{chao1965grammar}
Yuen~Ren Chao. 1965.
\newblock \emph{A grammar of spoken Chinese}.
\newblock Univ of California Press.

\bibitem[{Chen et~al.(2018)Chen, van Deemter, and Lin}]{chen2018modelling}
Guanyi Chen, Kees van Deemter, and Chenghua Lin. 2018.
\newblock Modelling pro-drop with the rational speech acts model.
\newblock In \emph{Proceedings of the 11th International Conference on Natural
  Language Generation}, pages 57--66. Association for Computational Linguistics
  (ACL).

\bibitem[{Dale(1989)}]{dale1989cooking}
Robert Dale. 1989.
\newblock Cooking up referring expressions.
\newblock In \emph{27th Annual Meeting of the association for Computational
  Linguistics}, pages 68--75.

\bibitem[{Dale(1992)}]{dale1992generating}
Robert Dale. 1992.
\newblock \emph{Generating referring expressions: Constructing descriptions in
  a domain of objects and processes.}
\newblock The MIT Press.

\bibitem[{Dale and Reiter(1995)}]{dale1995computational}
Robert Dale and Ehud Reiter. 1995.
\newblock Computational interpretations of the gricean maxims in the generation
  of referring expressions.
\newblock \emph{Cognitive science}, 19(2):233--263.

\bibitem[{Dale and Viethen(2009)}]{dale2009referring}
Robert Dale and Jette Viethen. 2009.
\newblock Referring expression generation through attribute-based heuristics.
\newblock In \emph{Proceedings of the 12th European workshop on natural
  language generation (ENLG 2009)}, pages 58--65.

\bibitem[{van Deemter(2016)}]{van2016computational}
Kees van Deemter. 2016.
\newblock \emph{Computational models of referring: a study in cognitive
  science}.
\newblock MIT Press.

\bibitem[{van Deemter and Gatt(2007)}]{Deemter2007Content}
Kees van Deemter and Albert Gatt. 2007.
\newblock Content determination in {GRE}: Evaluating the evaluator.
\newblock In \emph{Using Corpora for Natural Language Generation: Language
  Generation and Machine Translation}.

\bibitem[{van Deemter et~al.(2012)van Deemter, Gatt, Sluis, and
  Power}]{deemter2012generation}
Kees van Deemter, Albert Gatt, Ielka van~der Sluis, and Richard Power. 2012.
\newblock Generation of referring expressions: Assessing the incremental
  algorithm.
\newblock \emph{Cognitive science}, 36(5):799--836.

\bibitem[{van Deemter et~al.(2017)van Deemter, Sun, Sybesma, Li, Chen, and
  Yang}]{van-deemter-etal-2017-investigating}
Kees van Deemter, Le~Sun, Rint Sybesma, Xiao Li, Bo~Chen, and Muyun Yang. 2017.
\newblock \href {https://doi.org/10.18653/v1/W17-3532} {Investigating the
  content and form of referring expressions in {M}andarin: introducing the
  mtuna corpus}.
\newblock In \emph{Proceedings of the 10th International Conference on Natural
  Language Generation}, pages 213--217, Santiago de Compostela, Spain.
  Association for Computational Linguistics.

\bibitem[{Degen et~al.(2020)Degen, Hawkins, Graf, Kreiss, and
  Goodman}]{degen2020redundancy}
Judith Degen, Robert~D Hawkins, Caroline Graf, Elisa Kreiss, and Noah~D
  Goodman. 2020.
\newblock When redundancy is useful: A bayesian approach to
  “overinformative” referring expressions.
\newblock \emph{Psychological Review}.

\bibitem[{Dice(1945)}]{dice1945measures}
Lee~R Dice. 1945.
\newblock Measures of the amount of ecologic association between species.
\newblock \emph{Ecology}, 26(3):297--302.

\bibitem[{Fang et~al.(2015)Fang, Doering, and Chai}]{fang2015embodied}
Rui Fang, Malcolm Doering, and Joyce~Y Chai. 2015.
\newblock Embodied collaborative referring expression generation in situated
  human-robot interaction.
\newblock In \emph{Proceedings of the Tenth Annual ACM/IEEE International
  Conference on Human-Robot Interaction}, pages 271--278.

\bibitem[{Ferreira et~al.(2018)Ferreira, Moussallem, K{\'a}d{\'a}r, Wubben, and
  Krahmer}]{ferreira2018neuralreg}
Thiago~Castro Ferreira, Diego Moussallem, {\'A}kos K{\'a}d{\'a}r, Sander
  Wubben, and Emiel Krahmer. 2018.
\newblock Neuralreg: An end-to-end approach to referring expression generation.
\newblock \emph{arXiv preprint arXiv:1805.08093}.

\bibitem[{Gatt and Belz(2010)}]{gatt2010introducing}
Albert Gatt and Anja Belz. 2010.
\newblock Introducing shared tasks to nlg: The tuna shared task evaluation
  challenges.
\newblock In \emph{Empirical methods in natural language generation}, pages
  264--293. Springer.

\bibitem[{Gatt and Krahmer(2018)}]{gatt2018survey}
Albert Gatt and Emiel Krahmer. 2018.
\newblock Survey of the state of the art in natural language generation: Core
  tasks, applications and evaluation.
\newblock \emph{Journal of Artificial Intelligence Research}, 61:65--170.

\bibitem[{Gatt et~al.(2013)Gatt, Krahmer, van Gompel, and van
  Deemter}]{gatt2013production}
Albert Gatt, Emiel Krahmer, Roger van Gompel, and Kees van Deemter. 2013.
\newblock Production of referring expressions: Preference trumps
  discrimination.
\newblock In \emph{Proceedings of the Annual Meeting of the Cognitive Science
  Society}, volume~35.

\bibitem[{Gatt et~al.(2007)Gatt, van~der Sluis, and van
  Deemter}]{gatt2007evaluating}
Albert Gatt, Ielka van~der Sluis, and Kees van Deemter. 2007.
\newblock \href {http://staff.um.edu.mt/albert.gatt/pubs/enlg2007.pdf}
  {Evaluating algorithms for the generation of referring expressions using a
  balanced corpus}.
\newblock In \emph{Proceedings of the 11th European Workshop on Natural
  Language Generation (ENLG'07)}, pages 49--56, Schloss Dagstuhl, Germany.
  Association for Computational Linguistics.

\bibitem[{Gatt et~al.(2008)Gatt, van~der Sluis, and van Deemter}]{gatt2008xml}
Albert Gatt, Ielka van~der Sluis, and Kees van Deemter. 2008.
\newblock \href {http://www.csd.abdn.ac.uk/~ agatt/home/pubs/tunaFormat.pdf}
  {Xml format guidelines for the tuna corpus}.
\newblock Technical report, Technical report, Dept of Computing Science,
  University of Aberdeen.

\bibitem[{van Gompel et~al.(2019)van Gompel, van Deemter, Gatt, Snoeren, and
  Krahmer}]{van2019conceptualization}
Roger van Gompel, Kees van Deemter, Albert Gatt, Rick Snoeren, and Emiel~J
  Krahmer. 2019.
\newblock Conceptualization in reference production: Probabilistic modeling and
  experimental testing.
\newblock \emph{Psychological review}, 126(3):345.

\bibitem[{Gupta and Stent(2005)}]{gupta2005automatic}
Surabhi Gupta and Amanda Stent. 2005.
\newblock Automatic evaluation of referring expression generation using
  corpora.
\newblock In \emph{Proceedings of the Workshop on Using Corpora for Natural
  Language Generation}, pages 1--6. Citeseer.

\bibitem[{Horton and Keysar(1996)}]{horton1996speakers}
William~S Horton and Boaz Keysar. 1996.
\newblock When do speakers take into account common ground?
\newblock \emph{Cognition}, 59(1):91--117.

\bibitem[{Howcroft et~al.(2017)Howcroft, Vogels, and Demberg}]{howcroft2017g}
David~M Howcroft, Jorrig Vogels, and Vera Demberg. 2017.
\newblock G-tuna: a corpus of referring expressions in german, including
  duration information.
\newblock In \emph{Proceedings of the 10th International Conference on Natural
  Language Generation}, pages 149--153.

\bibitem[{Huang(1984)}]{huang1984distribution}
C-T~James Huang. 1984.
\newblock On the distribution and reference of empty pronouns.
\newblock \emph{Linguistic inquiry}, pages 531--574.

\bibitem[{James et~al.(2009)James, Li, and Li}]{james2009syntax}
Huang C-T James, Y-H~Audrey Li, and Yafei Li. 2009.
\newblock The syntax of chinese.
\newblock \emph{Cambridge, Cambridge). doi}, 10.

\bibitem[{Jordan and Walker(2005)}]{jordan2005learning}
Pamela~W Jordan and Marilyn~A Walker. 2005.
\newblock Learning content selection rules for generating object descriptions
  in dialogue.
\newblock \emph{Journal of Artificial Intelligence Research}, 24:157--194.

\bibitem[{Koolen et~al.(2011)Koolen, Gatt, Goudbeek, and
  Krahmer}]{koolen2011factors}
Ruud Koolen, Albert Gatt, Martijn Goudbeek, and Emiel Krahmer. 2011.
\newblock Factors causing overspecification in definite descriptions.
\newblock \emph{Journal of Pragmatics}, 43(13):3231--3250.

\bibitem[{Koolen and Krahmer(2010)}]{koolen2010d}
Ruud Koolen and Emiel Krahmer. 2010.
\newblock The d-tuna corpus: A dutch dataset for the evaluation of referring
  expression generation algorithms.
\newblock In \emph{LREC}.

\bibitem[{Krahmer and van Deemter(2012)}]{krahmer2012computational}
Emiel Krahmer and Kees van Deemter. 2012.
\newblock Computational generation of referring expressions: A survey.
\newblock \emph{Computational Linguistics}, 38(1):173--218.

\bibitem[{Levelt(1993)}]{levelt1993speaking}
Willem~JM Levelt. 1993.
\newblock \emph{Speaking: From intention to articulation}, volume~1.
\newblock MIT press.

\bibitem[{Lv(1979)}]{lv1979problems}
Shuxiang Lv. 1979.
\newblock Problems in the analysis of chinese grammar.

\bibitem[{Mao et~al.(2016)Mao, Huang, Toshev, Camburu, Yuille, and
  Murphy}]{mao2016generation}
Junhua Mao, Jonathan Huang, Alexander Toshev, Oana Camburu, Alan~L Yuille, and
  Kevin Murphy. 2016.
\newblock Generation and comprehension of unambiguous object descriptions.
\newblock In \emph{Proceedings of the IEEE conference on computer vision and
  pattern recognition}, pages 11--20.

\bibitem[{Newnham(1971)}]{newnham1971about}
Richard Newnham. 1971.
\newblock \emph{About {C}hinese}.
\newblock Penguin Books Ltd.

\bibitem[{Reiter and Dale(2000)}]{reiter2000building}
Ehud Reiter and Robert Dale. 2000.
\newblock \emph{Building natural language generation systems}.
\newblock Cambridge university press.

\bibitem[{van~der Sluis et~al.(2006)van~der Sluis, Gatt, and van
  Deemter}]{van2006manual}
Ielka van~der Sluis, Albert Gatt, and Kees van Deemter. 2006.
\newblock \href
  {http://homepages.abdn.ac.uk/k.vdeemter/pages/TunaCorpusManual/index.html}
  {Manual for the tuna corpus: Referring expressions in two domains}.
\newblock Technical Report AUCS/TR0705, Department of Computing Science, Univ.
  of Aberdeen.

\bibitem[{van~der Sluis et~al.(2007)van~der Sluis, Gatt, and van
  Deemter}]{van2007evaluating}
Ielka van~der Sluis, Albert Gatt, and Kees van Deemter. 2007.
\newblock \href {http://staff.um.edu.mt/albert.gatt/pubs/ranlp2007.pdf}
  {Evaluating algorithms for the generation of referring expressions: Going
  beyond toy domains}.
\newblock In \emph{Proceedings of the International Conference on Recent
  Advances in Natural Language Processing (RANLP'07)}, Borovets, Bulgaria.
  RANLP.

\bibitem[{Yi et~al.(2020)Yi, Deng, and Hu}]{yi-etal-2020-improving}
Yanzhi Yi, Hangyu Deng, and Jinglu Hu. 2020.
\newblock \href {https://doi.org/10.18653/v1/2020.acl-main.93} {Improving image
  captioning evaluation by considering inter references variance}.
\newblock In \emph{Proceedings of the 58th Annual Meeting of the Association
  for Computational Linguistics}, pages 985--994, Online. Association for
  Computational Linguistics.

\end{thebibliography}
\bibliographystyle{acl_natbib}

\appendix
\section{Annotating \textsc{mtuna}}

Following and adapting from \citet{van2006manual}, the attributes and corresponding values we used in our annotation (annotating the \textsc{mtuna} and re-annotating the \textsc{etuna}) are shown in Table~\ref{tab:furniture} and~\ref{tab:people} for furniture and people domain respectively. An example of the annotated RE sample in \textsc{mtuna} is shown in Figure~\ref{fig:annotation}.

\begin{table}[!ht]
\begin{center}
  \begin{tabular}{ccc}
    \toprule
    Attribute & Possible Values & Freq.\\
    \midrule
    \texttt{TYPE} & \emph{chair, sofa, desk, fan} & 347\\
    \texttt{COLOUR} & \emph{blue, red, green, grey} & 326 \\
    \texttt{ORIENTATION} & \emph{front, back, left, right} & 185 \\
    \texttt{SIZE} & \emph{large, small} & 141 \\
    \texttt{X-DIMENSION} & 1, 2, 3, 4, 5 & 5\\
    \texttt{Y-DIMENSION} & 1, 2, 3 & 5\\
    \texttt{OTHER} & - & 10\\
    \bottomrule
  \end{tabular}
  \caption{Attributes and their values for REs in furniture domain.}
  \label{tab:furniture}
\end{center}

\end{table}

\begin{table}[!ht]
\begin{center}
  \begin{tabular}{ccc}
    \toprule
    Attribute & Possible Values & Freq.\\
    \midrule
    \texttt{TYPE} & \emph{person}& 278 \\
    \texttt{AGE} & \emph{young, old} & 74\\
    \texttt{ORIENTATION} & \emph{front, left, right} & 6\\
    \texttt{hasBread} & 0, 1 & 169\\
    \texttt{breadColour} & \emph{dark, light} & 95\\
    \texttt{hasHair} & 0, 1 & 116\\
    \texttt{hairColour} & \emph{dark, light} & 100\\
    \texttt{hasGlasses} & 0, 1 & 165 \\
    \texttt{hasShirt} & 0, 1 & 14\\
    \texttt{hasTie} & 0, 1 & 26\\
    \texttt{hasSuit} & 0, 1 & 60\\
    \texttt{X-DIMENSION} & 1, 2, 3, 4, 5 & 6\\
    \texttt{Y-DIMENSION} & 1, 2, 3& 6\\
    \texttt{OTHER} & - & 151 \\
    \bottomrule
  \end{tabular}
  \caption{Attributes and their values for REs in people domain.}
  \label{tab:people}
\end{center}

\end{table}

\begin{figure}[h]
\begin{lstlisting}[frame=single, escapeinside=``, basicstyle=\ttfamily\footnotesize]
{ "sno": "Object", 
  "subject_id": "2", 
  "object": [
    { "attributes": [
        {"name": "COLOUR", 
          "value": "dark"
        }, 
        {"name": "TYPE", 
          "value": "table"
        }]
    }], 
  "trial_id": "1", 
  "utt": "`灰桌子`"
}
\end{lstlisting}
\caption{An example annotated data sample from \textsc{mtuna}, for the RE \emph{灰桌子} (\emph{grey table}).}
\label{fig:annotation}
\end{figure}

\section{Results for Annotating Over-/Under-specifications}

Table~\ref{tab:scene} listed all trails we used for analysing the use of over- and under-specifications in this study, each of which we annotated how many alternative minimal descriptions and numerical over-specifications are there. The raw data of the \textsc{mtuna} corpus (and trial ID) can be found at: \url{homepages.abdn.ac.uk/k.vdeemter-/pages/mtuna-webpage/}, and those of \textsc{etuna} are available at: \url{https://www.abdn.ac.uk/ncs/departments/computing-science/tuna-318.php}.

\begin{table}[!ht]
\centering
\begin{tabular}{cccc}
    \toprule
    \textbf{\textsc{mtuna}} & \textbf{\textsc{etuna}} & \textbf{\#mini.} & \textbf{\#num.} \\
    \midrule
    1 & 1 & 1 & 0 \\
    2 & 2 & 1 & 0 \\
    3 & 3 & 1 & 0 \\
    4 & 4 & 1 & 0 \\
    5 & 5 & 1 & 0 \\
    6 & 6 & 1 & 0 \\
    7 & 7 & 1 & 0 \\
    8 & - & 1 & 1 \\
    9 & - & 2 & 0 \\
    10 & - & 3 & 0 \\
    \midrule
    21 & 21 & 1 & 1 \\
    22 & 22 & 1 & 0 \\
    23 & 23 & 1 & 0 \\
    24 & 24 & 1 & 0 \\
    25 & 25 & 2 & 1 \\
    26 & 26 & 1 & 2 \\
    27 & - & 1 & 2 \\
    28 & - & 1 & 2 \\
    29 & - & 2 & 1 \\
    30 & - & 1 & 3 \\
    \bottomrule
\end{tabular}
\caption{The number of possible minimal descriptions (\textbf{mini.} and numerical over-specifications (\textbf{num.}) for each trial. The first two colomns are the trial IDs in \textsc{mtuna} and \textsc{etuna}.}
\label{tab:scene}
\end{table}

\end{CJK}

\end{document}